\documentclass{article}
\usepackage{spconf,amsmath,graphicx}
\usepackage[usenames,dvipsnames]{color} 

\title{Multispectral image fusion based on super pixel segmentation}
\name{Nati Ofir and Jean-Christophe Nebel}

\address{Kingston University London}
\begin{document}
	%
	\maketitle

	\begin{abstract}
        Multispectral image fusion is a computer vision process that is essential to remote sensing. For applications such as dehazing and object detection,
        there is a need to offer solutions that can perform in real-time on any type of scene. Unfortunately, current state-of-the-art approaches do not meet these criteria as they need to be trained on domain-specific data and have high computational complexity.
        This paper focuses on the task of fusing color (RGB) and near-infrared (NIR) images as this the typical RGBT sensors, as in multispectral cameras for detection, fusion, and dehazing.
        Indeed, the NIR channel has the ability to capture details not visible in RGB and see beyond haze, fog, and clouds. 
        To combine this information, a novel approach based on superpixel segmentation is designed so that multispectral image fusion is performed according to the specific local content of the images to be fused. 
        Therefore, the proposed method produces a fusion that contains the most relevant content of each spectrum. 
        The experiments reported in this manuscript show that the novel approach 
        better preserve details than alternative fusion methods. 
        \end{abstract}
	
\section{Introduction} \label{sec:intro}

\begin{figure}[tbh]
	\centering
	\includegraphics[width=78px]{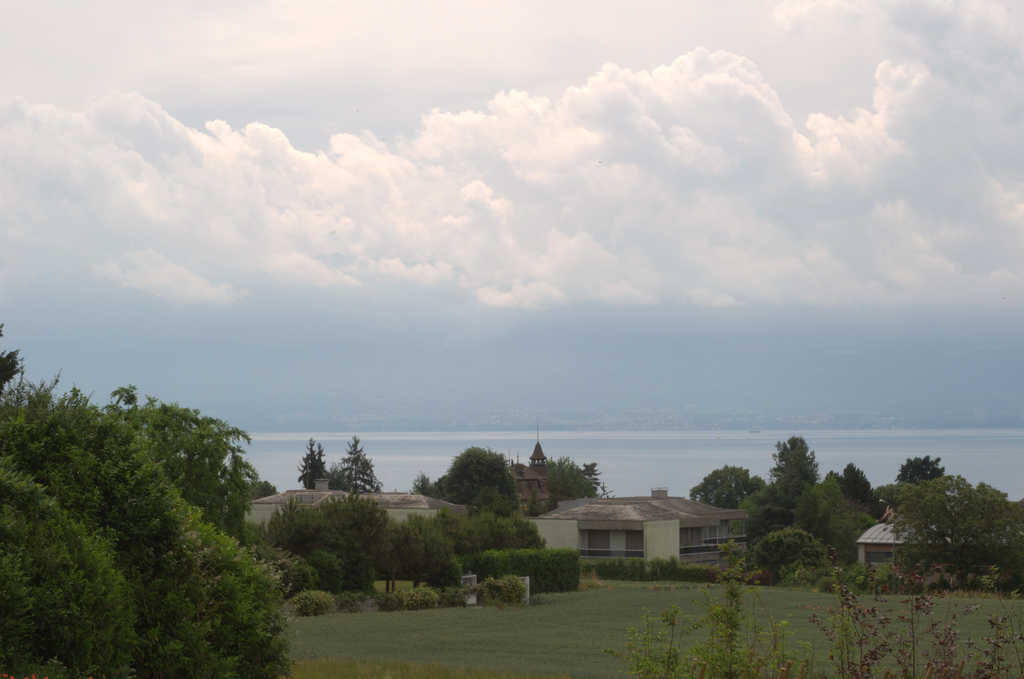}
	\includegraphics[width=78px]{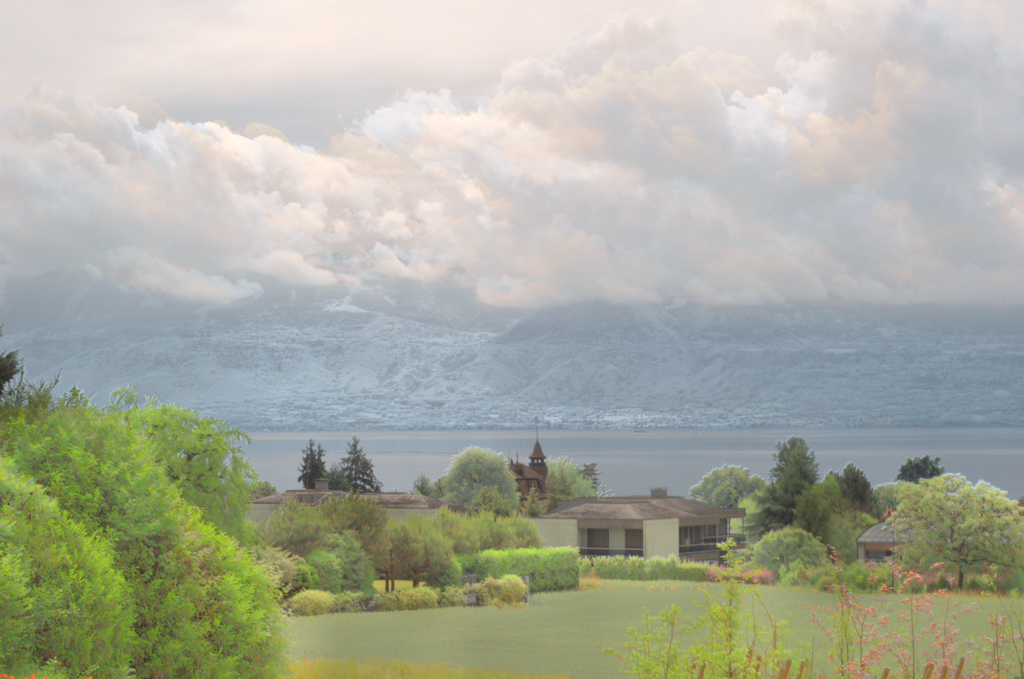}
	\includegraphics[width=78px]{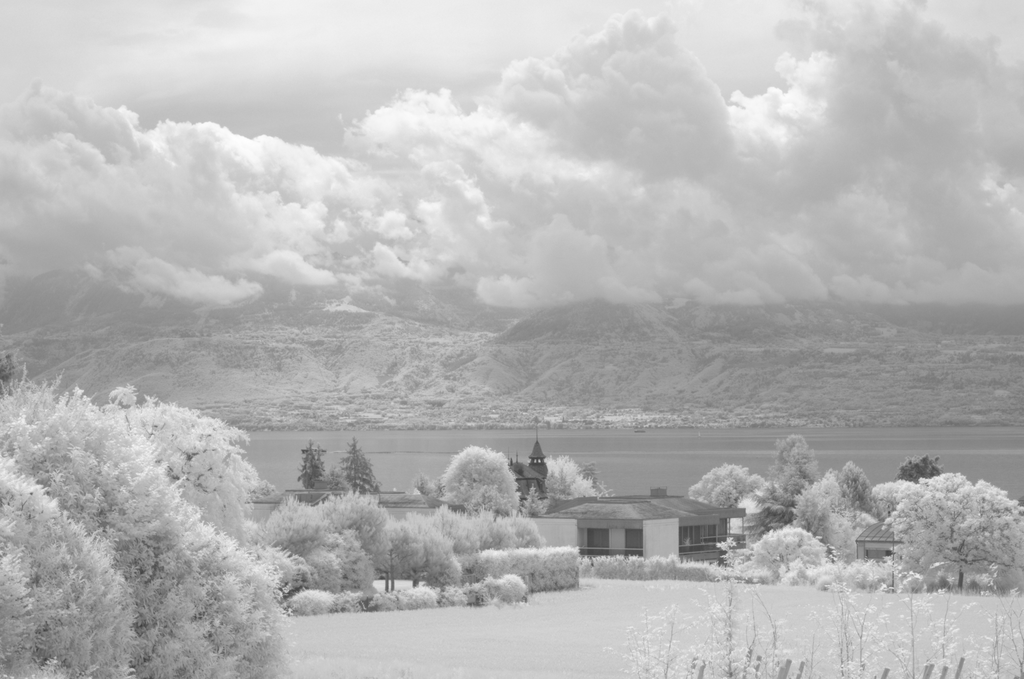}
	\caption{Example of the proposed fusion method results. From left to right: input RGB image, the proposed fusion result, input NIR image. The proposed fusion contains both the color information of the RGB and the far details captured by the NIR image.}
	\label{fig:fusion}
\end{figure}

Image fusion is an important task of image processing that has numerous applications such as detection, dehazing, and visualization. It is relying on modern and defense cameras and has challenged the research communities for a couple of decades. This manuscript introduces a new approach to fuse multispectral images based on the content of regions defined through superpixel segmentation. As this research is focused on multispectral images for object detection \cite{takumi2017multispectral}, dehazing \cite{guo2020rsdehazenet}, and fusion visualization like in satellite imaging \cite{zeng2010fusion}, we specifically address the data of the RGB-NIR dataset of \cite{BS11}, however, it can be extended easily to other modalities.
Therefore, this work focuses on the fusion of RGB visible color $0.4-0.7\mu m$ and NIR $0.7-2.5\mu m$ images as described in \cite{ofir2018deep}. Since each spectrum captures different information about a scene, their fusion is challenging and informative. While the RGB channel captures the visible color, as seen by human eyes, the NIR channel can see beyond haze, fog, and clouds, and thus can in particular reveal far-distance details in a scene. Figure \ref{fig:fusion} illustrates scene enhancements that the proposed multispectral image fusion delivers. Indeed, the fused image contains both color information and the structure of the far mountains that were invisible in the original RGB image.

As more and more sensors are integrated into modern cameras, the ability to fuse effectively their inputs in real-time is essential to produce the most informative pictures. A non-trivial preprocessing step \cite{lowe2004distinctive}, which is out of the scope of this work, is the alignment of the sensor data based on multispectral image registration \cite{ofir2018registration,ofir2018deep}. 

Once aligned, the fusion of the input images can be performed using a variety of approaches including $\alpha$-blending, Principal-Component-Analysis (PCA) blending \cite{he2010multimodal} and spectral blending like Wavelet fusion \cite{chipman1995wavelets}. Although deep neural network approaches have recently been proposed \cite{chen2017deep}, they require large databases and heavy computational resources that are often unavailable in real-life projects with unique spectra, such as object detection of autonomous driving, of fusing multispectral images captured from a satellite.
In this paper, the focus is on real-time and generic approaches for multispectral image fusion, while the algorithms behind PCA and spectral fusions are detailed as they have inspired the proposed solution. Then, it describes our approach which is spatial and based on superpixel segmentation of the input images. 

We apply a soft mask fusion such that the blending weight is changing between every pixel in the image. This soft mask emphasizes the details in the fused images.

This manuscript is organized as follows. After a review of relevant work in Section \ref{sec:previous}, Section \ref{sec:review} presents further details on existing approaches that have inspired the proposed solution and are used for its evaluation. Then Section \ref{sec:fusion} describes the new spatial approach for image fusion that relies on superpixel content. Next, Section \ref{sec:results} reports both qualitative and quantitative results demonstrating the added value of the proposed fusion approach. Finally, conclusions and
future work are presented in Section \ref{sec:conclusions}.

\section{Previous Work} \label{sec:previous}

Multispectral image registration 
and fusion is a challenging task that has been studied in the last decades. It has been addressed by both traditional \cite{ofir2018registration} and deep learning (DL)-based \cite{ofir2018deep} computer vision approaches. 
Whereas DL solutions tend to deliver much better performance in the specific domain for which they were trained, traditional algorithms are generally rooted in stronger theoretical frameworks, have lower computational complexity, are more generic, and do not rely on the availability of suitable training data \cite{ofir2021classic,ofir2021classicThesis}.
Image fusion has an application in multispectral images \cite{wei2015hyperspectral} and in medical imaging \cite{james2014medical}. Image fusion is also important in processing images from a single modality as in multi-focus \cite{zhao2018multi}. This work is addressing specifically the problem of multispectral image fusion.

Image fusion can be addressed through various technical approaches. Toet et al \cite{toet1990hierarchical} use a hierarchical approach of fusion. A novel group of works introduced fusion according to the Principal Component Analysis (PCA) \cite{kumar2006pca,srivastava2017combination}. Early methods relied on global statistics \cite{he2010multimodal} or on spectral properties like in wavelet fusion \cite{pajares2004wavelet}. Advanced methods use guided filtering for fusion \cite{li2013image}.
Following the DL revolution in computer vision, such approaches have become the main sources of investigation for a fusion of images, features, and data. For example, Chen et al \cite{chen2017deep} exploited deep fusion to improve classification accuracy, while convolutional neural networks were successfully used to detect pedestrians \cite{wagner2016multispectral}. 
As these DL methods rely on large datasets and heavy training processes, they are not suitable for the applications targeted in this work. Still, it is worth mentioning that a recent DL solution took advantage of the concept of superpixel for hyperspectral fusion \cite{jia2019collaborative} incorporating texture information. 

The proposed method of this paper introduces a method that incorporates low-level statistics of image superpixels. Due to its low computational complexity, it could run in real-time on simple devices and embedded systems
and produces informative fused multispectral images. In contrast to previous works, we are focused on the multispectral case for multi-sensor cameras, astrophysics satellites and defense systems that are typically in the near to visible light frequencies such as UV and IR, and in addition, do not contain heavy computational resources since their systems are embedded. 

\section{Concepts behind Relevant Fusion Methods} \label{sec:review}

In this section, further details are provided about approaches that have inspired the proposed solution. They comprise global $\alpha$ blending, PCA, and spectral fusions.

\subsection{The $\alpha$ Blending Fusion} \label{sec:alpha}

Global $\alpha$ blending 
is a simple standard image fusion technique. Given multispectral image $NIR,RGB$ that are NIR and visible color images respectively we first convert them to grayscale such that:
$I_1 = NIR,I_2 = gray(RGB)$.
Then given a constant $\alpha \in [0,1]$ the $\alpha$ blending gray fusion is:
$F_{gray} = \alpha \cdot I_1+(1-\alpha) \cdot I_2$.
Finally, the colored fusion image is produced using the following color preservation formula:
$F = \frac{F_{gray}}{I_2} \cdot RGB$.

\subsection{PCA Fusion} \label{sec:pca}

The PCA fusion computes $\alpha$ according to the joint statistics of the fused images as follows. Given the gray images, $I_1,I_2$, their individual variance, 
$V_j = Var(I_j(x))$,
and their joint covariance,
$C = cov(I_1(x),I_2(x))$,
are calculated.
PCA decomposition is then applied by computing the eigenvalues, $\lambda$, and eigenvectors of their covariance matrix:
\begin{equation}
|\lambda I-\begin{pmatrix}
V_1 & C \\ C & V_2 
\end{pmatrix} | = (\lambda-V_1)(\lambda-V_2)-C^2
\end{equation}
Subsequently, by setting that this equation equals to zero,
$\lambda^2-\lambda(V_1+V_2)+V_1V_2-C^2 = 0$, 
its discriminant, $\Sigma$, can be computed: 
$\Sigma = \sqrt{(V_1+V_2)^2-4*V_1V_2-4C^2}$,
i.e., $\Sigma=\sqrt{V_1^2+V_2^2-2*V_1V_2-4C^2}$.

\noindent Next, the eigenvalues, $\lambda_{1,2}$ are evaluated:
$\lambda_{1,2} = \frac{V_1+V_2 \pm \Sigma}{2}$,    
leading to the eigenvector, $v_{\lambda_{1}}$, being expressed as:
\begin{equation}
v_{\lambda_{1}} = [1,\frac{\lambda_1-V_1}{C}]^T    
\end{equation}

\noindent Note that as it is assumed the first eigenvalue is the larger principal component, the second smaller eigenvalue is neglected.

\noindent Finally the blending factor is set as:
$\alpha_{PCA} = ||v_{\lambda_{1}}||^{-1}$.

To conclude, this process allows performing multispectral image fusion based on global PCA coefficients. In the experiment Section \ref{sec:results}, this approach is used as baseline to evaluate the proposed fusion method based on region content.

\subsection{Spectral Fusion} \label{sec:spectral}


Spectral methods has been an approach of choice to perform local image fusion \cite{ofir2021classic,pajares2004wavelet}. These methods rely on the computation of low and high pass filters. 
The low pass filter of an image, $I_j$, is computed by:
$LP_j = I_j * Gaussian(x,y)$, where $x,y$ are the spatial filter pixels.

Then, the high pass filter is calculated as follows:
$HP_j = I_j - LP_j$.
Then, each band-pass is processed using a different fusion mechanism. 

For the low frequencies, $\alpha$ blending is applied:
$F_{LP} = \alpha \cdot LP_1+(1-\alpha) \cdot LP_2$.
where $\alpha$ is typically set to 0.5 when no additional information is known about the images.

For the high frequencies, a maximum criterion is used:
$F_{HP}(x) = max(HP_1(x),HP_2(x))$.

Finally, the overall fusion is derived by adding the different frequencies:
$F_{gray} = F_{LP}+F_{HP}$.

Spectral fusion forms the basis for the proposed content-based fusion. Although, similar to spectral fusion, the region content-based fusion puts emphasis on high frequencies: it calculates the gain multiplier
according to the content of the local superpixels using pixel-wise soft maps.

\section{The Proposed Method: Super Pixel Region-based Fusion} \label{sec:fusion}

In this section, we introduce the proposed approach of multispectral image fusion by region content analysis. It extends the concepts behind PCA and spectral fusion by relying on the local content of the fused images that are expressed using high-order statistics to compute fusion weights for each pixel. Note that although the presented approach addresses the challenge of fusing two images, it could easily be extended to process any number of images as in hyperspectral imaging \cite{wei2015hyperspectral}.

The fusion of the two input images uses a soft smooth pixel map using a mask with pixel-wise fusion weights:
\begin{equation}
F_{gray}(x) = mask(x) \cdot I_1(x) + (1-mask(x)) \cdot I_2(x)
\end{equation}
This fusion is performed in a hierarchical manner using Laplacian Pyramids \cite{burt1987laplacian} to apply visual-informative masking with $mask(x)$.
In the next steps, the construction of a smooth $mask(x,y)$ is detailed. Note that $mask(x,y)$ is a function of the content in the fused images based on the superpixel region content segmentation.
Given an input image $I$ comprised of pixels, are $\{x_1,...,x_N\}$, the proposed method performs a superpixel decomposition \cite{achanta2012slic} of the pixels such that:
$X = \{x_1,...,x_N\} = \bigcup_j^k X_j$.

As the standard deviation of a superpixel has a significant correlation with its content level \cite{achanta2012slic}, for each superpixel $X_j$ of the input image as follows, its standard deviation is calculated by the expected mean $E$:
\begin{equation}
std(X_j) = \sqrt{E_{x \in X_j}[I^2(x)]-E^2_{x \in X_j}[I(x)]}
\end{equation}
Here the grade of a superpixel $X_j$ is defined by its standard deviation:
$grade(X_j) = std(X_j)$.

Eventually, the grade of a single pixel in each input image, i.e., its statistical score, is set to the value of the superpixel it belongs to:
$\forall x \in X_j: grade(x) = grade(X_j)$.

The final $mask(x)$ should maximize the grades of the fused images while maintaining global smoothness. Inspired by optical flow regularization \cite{horn1981determining}, this can be achieved by finding the $argmax$ of the following formula:
\begin{equation}
\sum mask\cdot grade(I_1) + (1-mask) \cdot grade(I_2)-\beta ||\bigtriangledown mask||^2
\end{equation}

According to Euler-Lagrange equation \cite{agrawal2002formulation}, and the general approximation suggested by \cite{horn1981determining}, a solution can be found using an iterative process:
\begin{equation}
mask_{j+1} = \hat{mask_j}+\frac{grade(I_1)-grade(I_2)}{4\beta},
\end{equation}
where $\hat{mask}$ is the four-pixel-neighbor average.

\begin{figure}[tbh]
	\centering
	\includegraphics[width=60px]{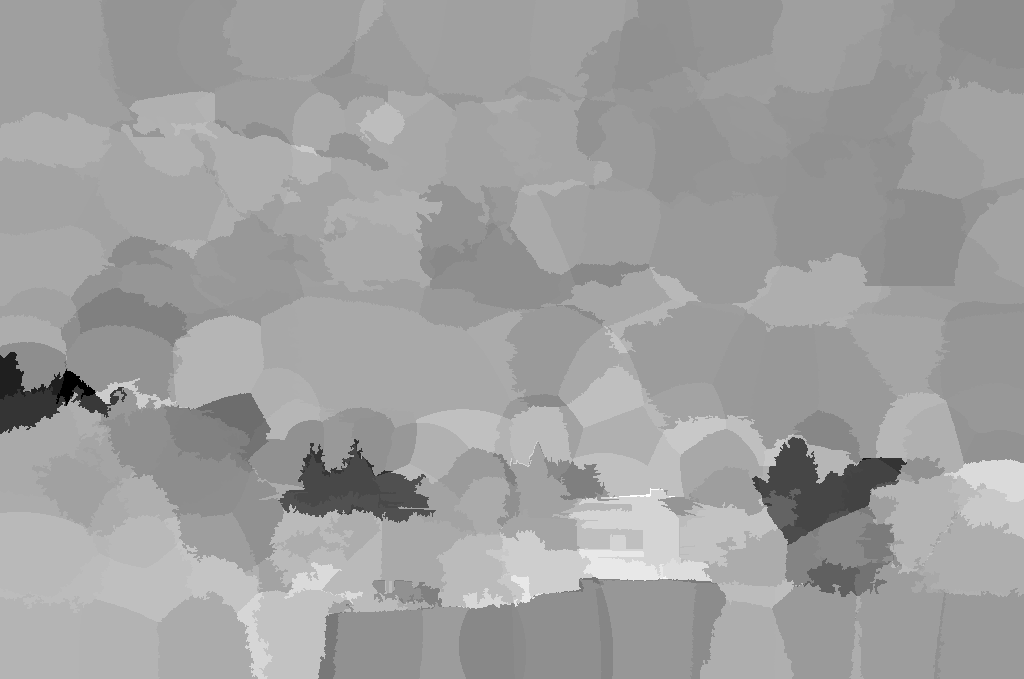}
	\caption{Mask of pixel blending computed by the proposed method for images shown in Figure \ref{fig:fusion}. Darker shades indicate higher weights for the pixels of the RGB image.}
	\label{fig:mask}
\end{figure}

It can be shown 
that a heuristic to find such a solution mask can be given by the difference of grades together with a smoothing operation on top of it:
\begin{equation}
mask(x) = \sigma[grade(I_1(x))-grade(I_2(x))]
\end{equation}
The sigmoid $\sigma(x) = \frac{e^x}{1+e^x}$ is used to emphasize the level information in the fused image pixels. Finally, $mask(x)$ is normalized linearly so that it lies in the range of $[0,1]$ and applies a Gaussian smoothing on its values. See Figure \ref{fig:mask} for an example of a fusion mask of the proposed method of superpixel fusion, where black and white pixels correspond to scores of 0 and 1, respectively.

Having described the concepts and algorithmic steps behind the proposed approach, the next section provides its evaluation on multispectral image fusion.

\section{Results} \label{sec:results}

The proposed approach is evaluated on the NIR-RGB multispectral dataset of \cite{BS11} since it contains natural images that can be captured by any typical camera both in the wild and in indoors.
This dataset contains aligned multispectral images, with 954 pairs, and it is divided into content types, such as country (104 images), mountain (110 images), and urban (116 images).

To assess that the produced fused images are meaningful, i.e., contain the color information of the RGB together with the far details of the NIR images, they are evaluated both quantitatively, i.e, by edge map preservation and ssim scores \cite{ssim, sara2019image},
and qualitatively, i.e., through comparison with the outcomes of the well-known PCA fusion \cite{wang2009mean} and spectral fusion \cite{ofir2018registration}. 
Note that all the compared methods are of linear complexity delivering runtime results in a few milliseconds. 
Quantitative evaluation is performed by measuring edge preservation of the multispectral image fusion as this  estimates how much of the edge content and details of the multispectral images remain in the fused images. 

By defining $C_1,C_2$ as the Canny \cite{canny1986computational} edge maps of the input images, and $C_F$ as the edge map of the fused image, the edge preservation metric is expressed by $\frac{1}{2}\sum_i\frac{\sum C_i \cdot C_F}{\sum C_i}$.

\begin{figure}[tbh]
	\centering
	\includegraphics[width=50px]{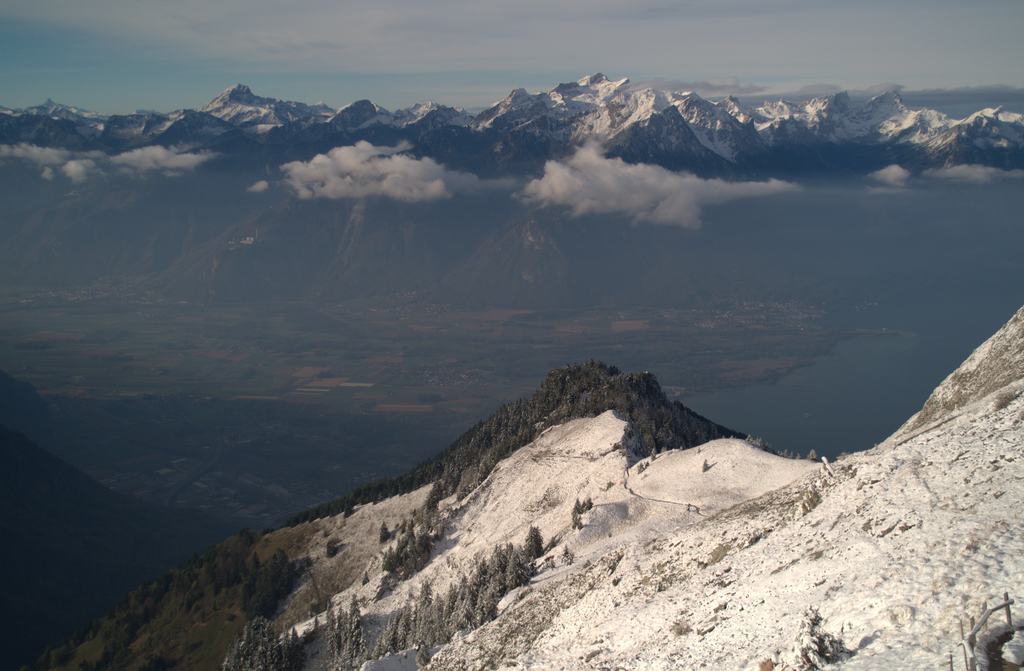}
	\includegraphics[width=50px]{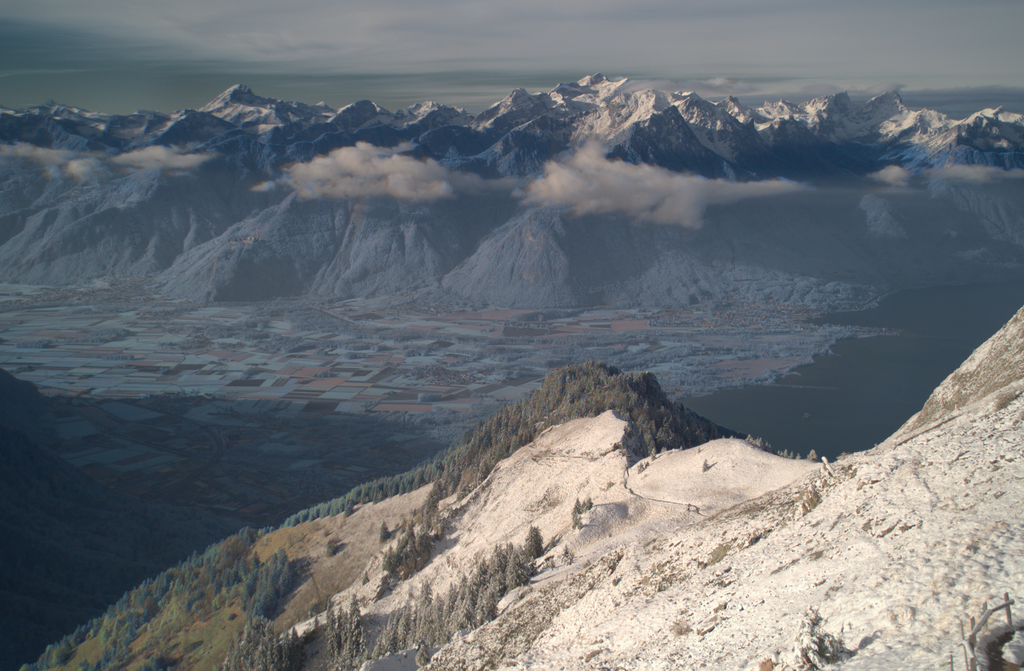}
	\includegraphics[width=50px]{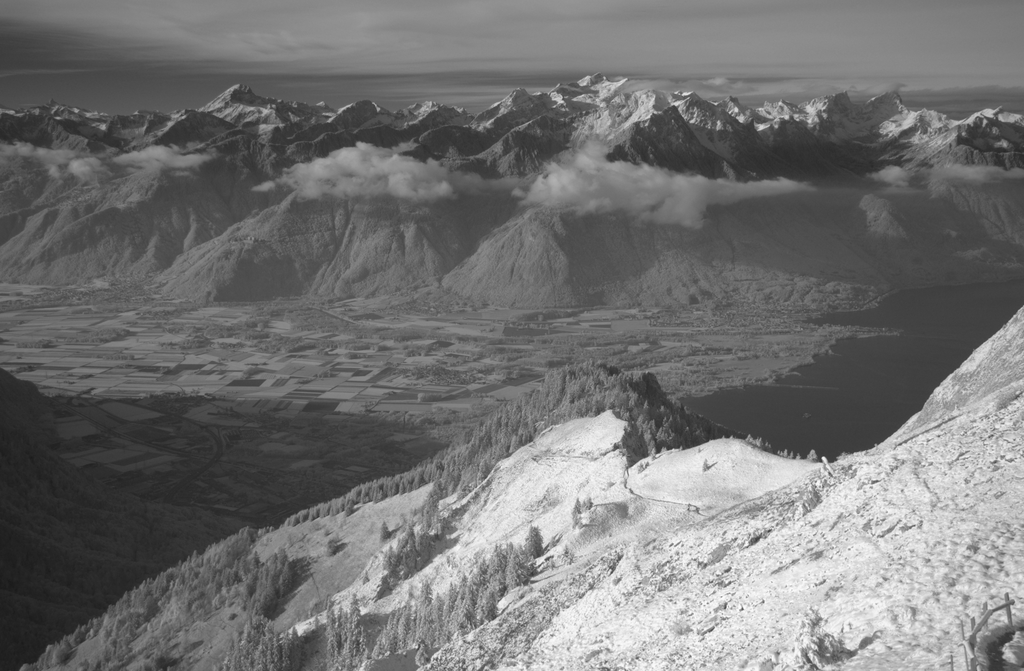}\\ [0.1cm]
	\includegraphics[width=50px]{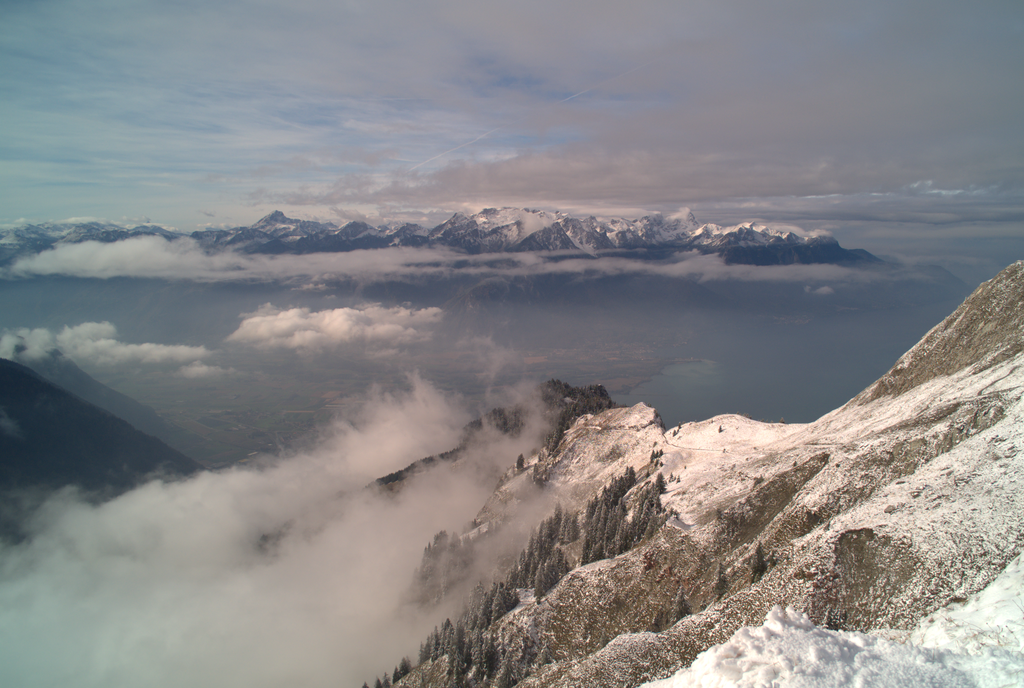}
	\includegraphics[width=50px]{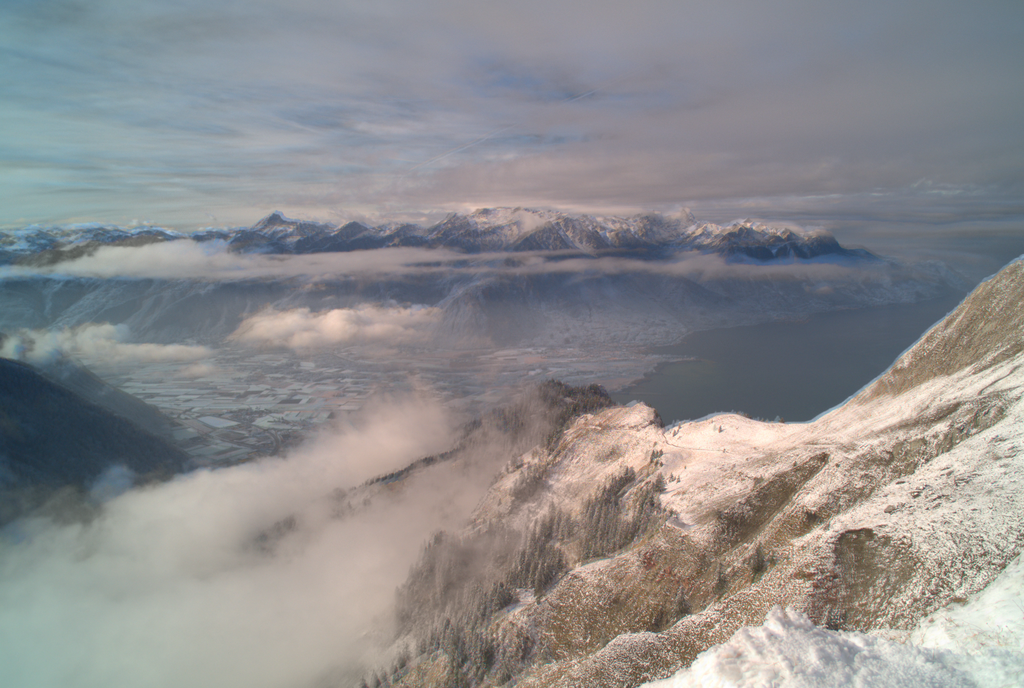}
	\includegraphics[width=50px]{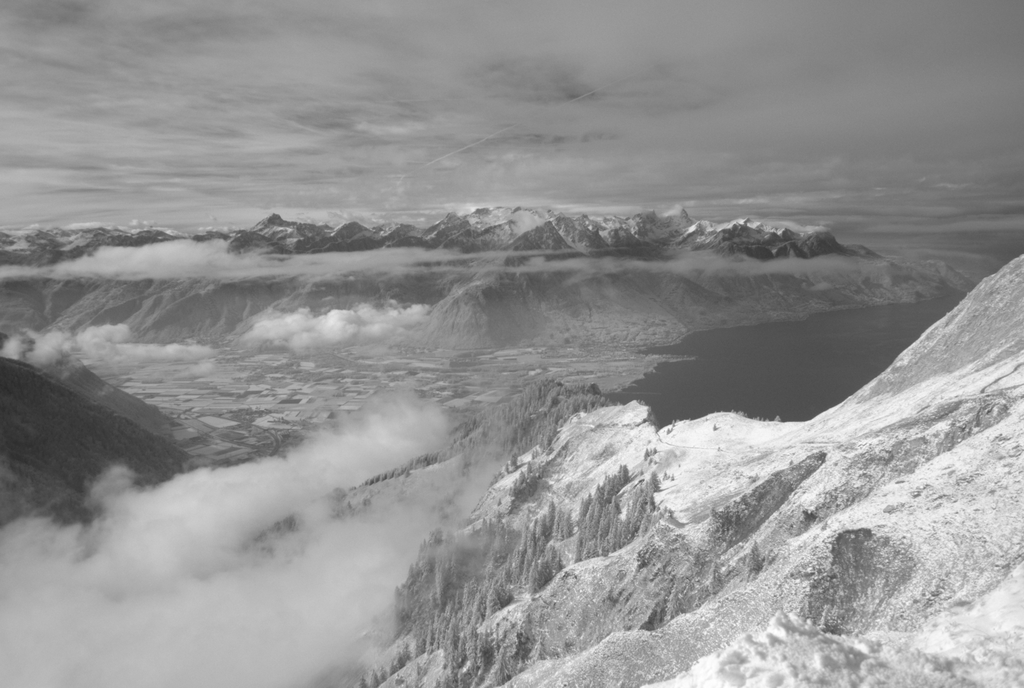}\\ [0.1cm]
	\includegraphics[width=50px]{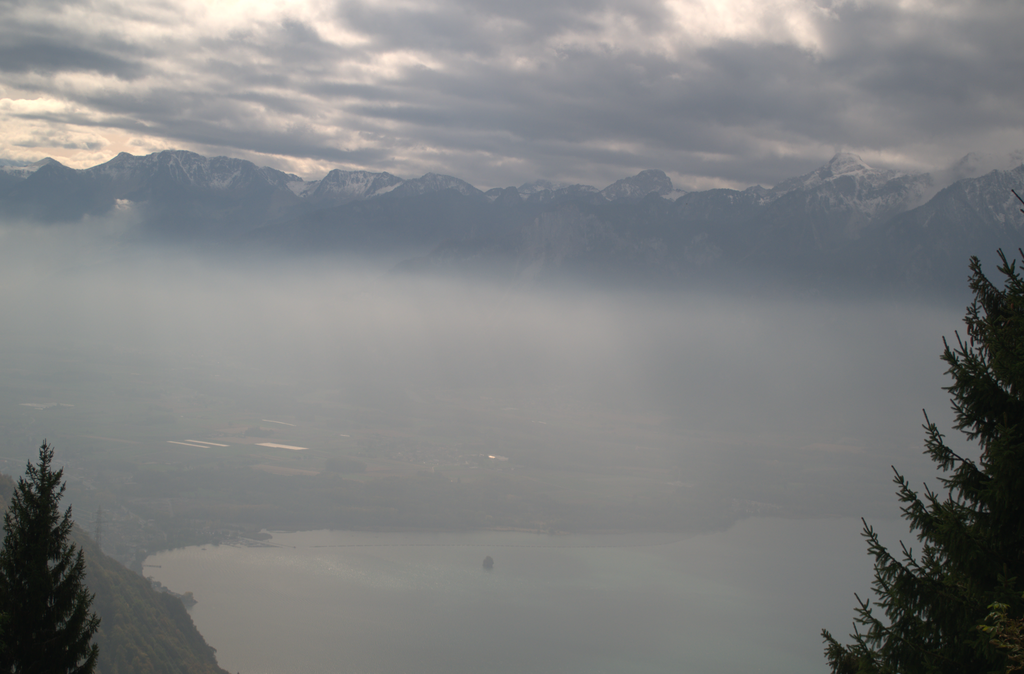}
	\includegraphics[width=50px]{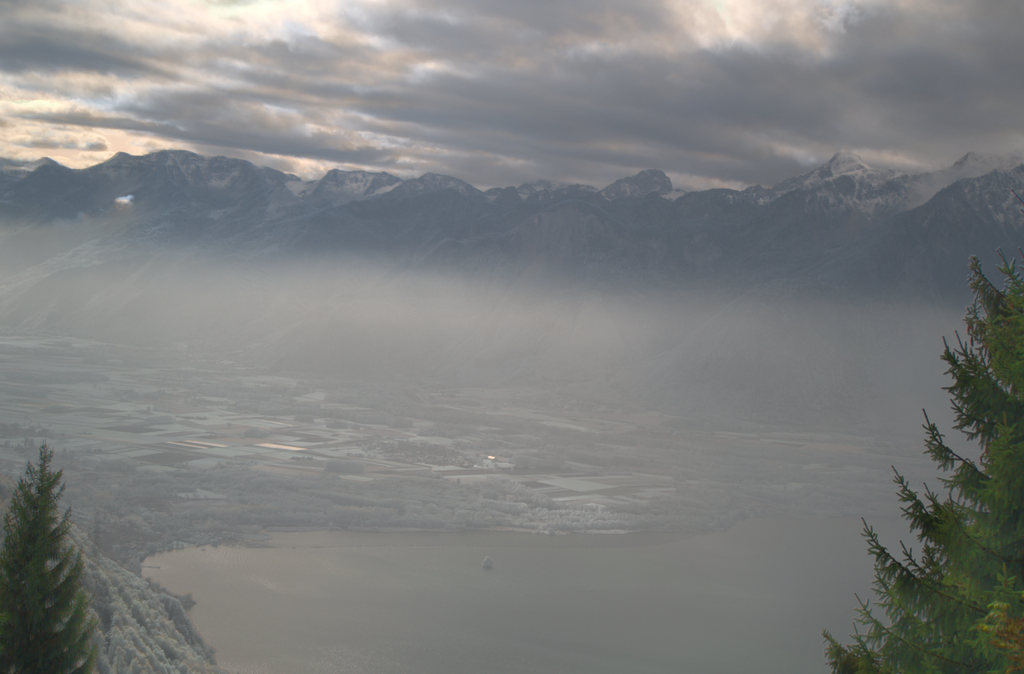}
	\includegraphics[width=50px]{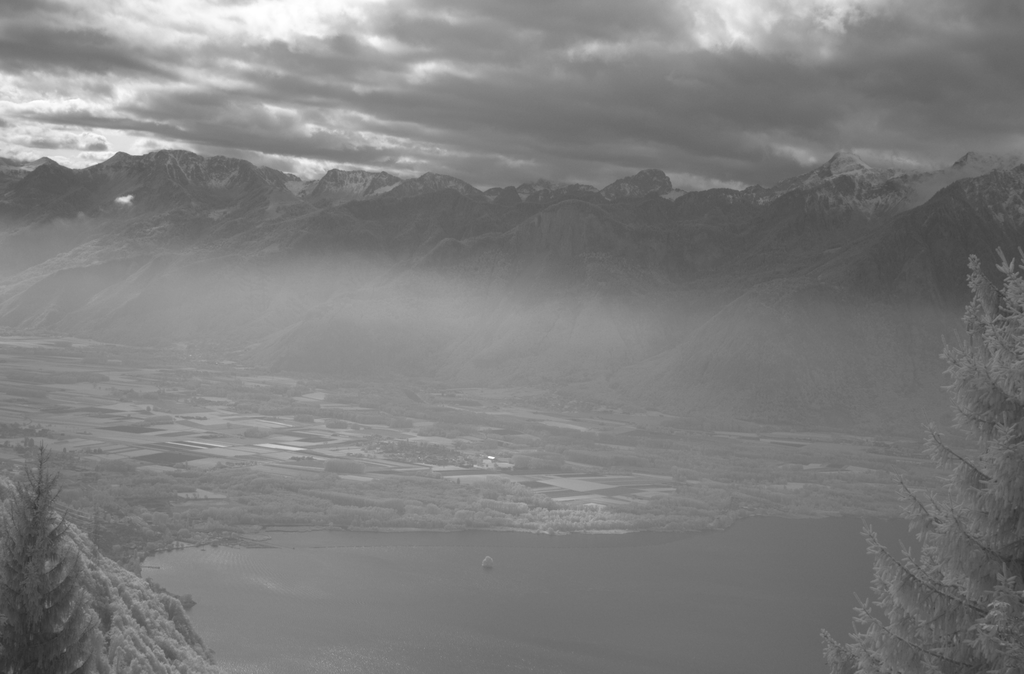}\\ [0.1cm]
	\includegraphics[width=50px]{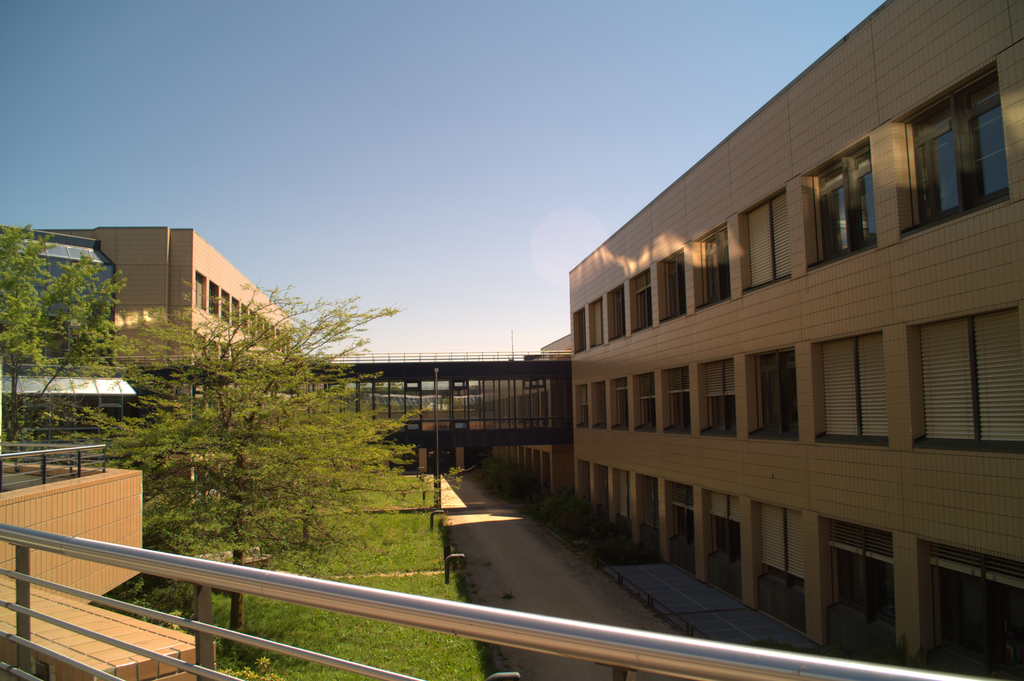}
	\includegraphics[width=50px]{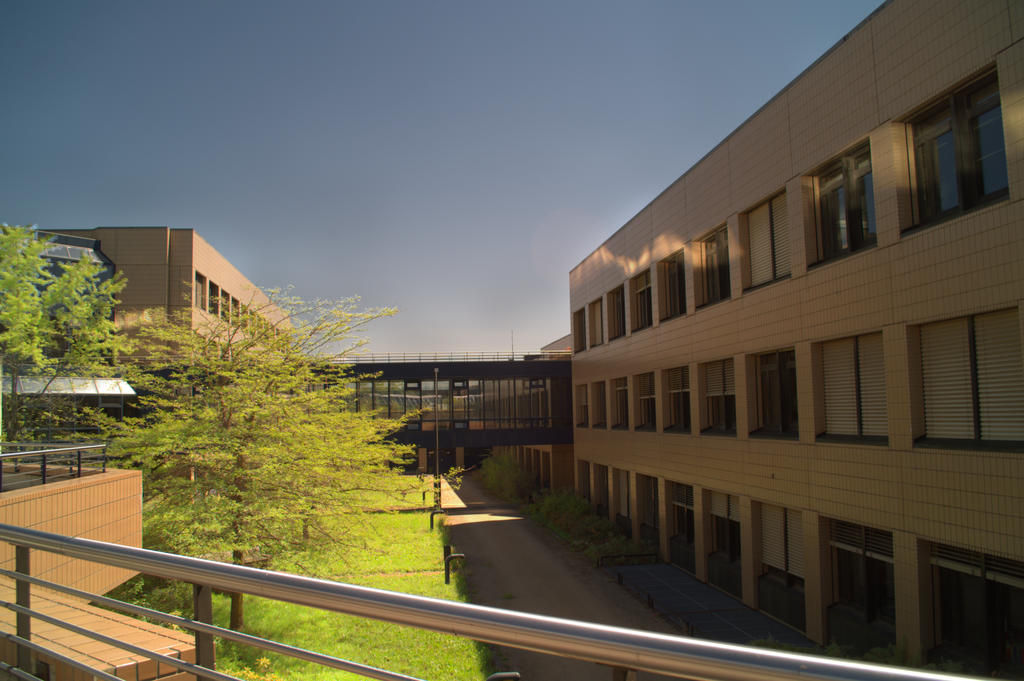}
	\includegraphics[width=50px]{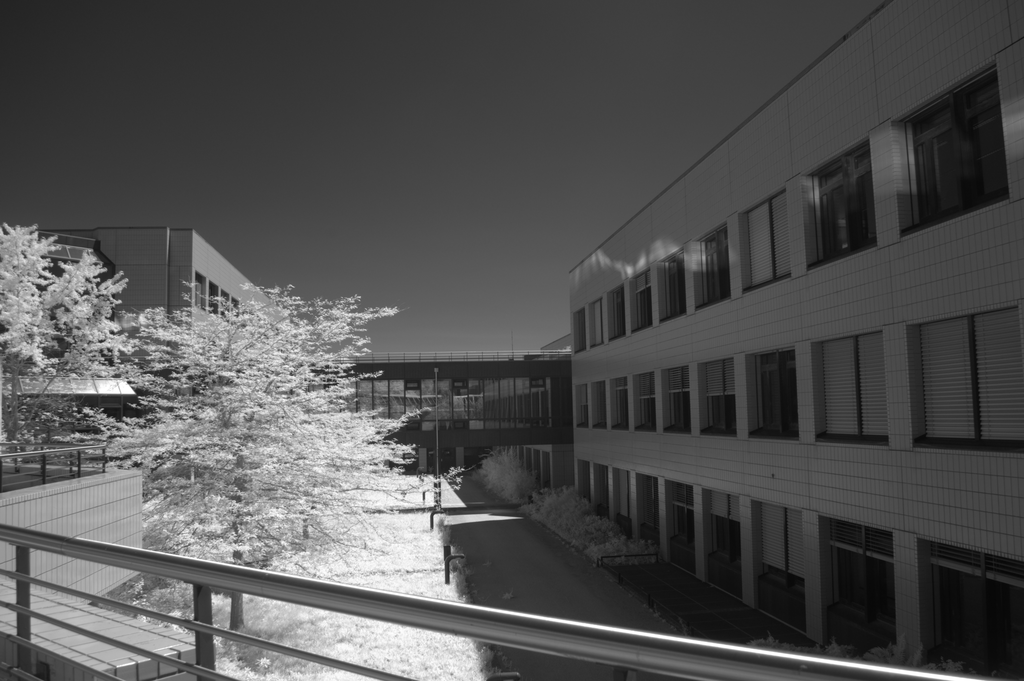} 
	\caption{Outcomes of the proposed multispectral image fusion. From left to right: input RGB,  fused, and input NIR images.}
	\label{fig:fusion_results}
\end{figure}

\begin{figure}[tbh]
	\centering
	\includegraphics[width=50px]{1_fusion.png}
	\includegraphics[width=50px]{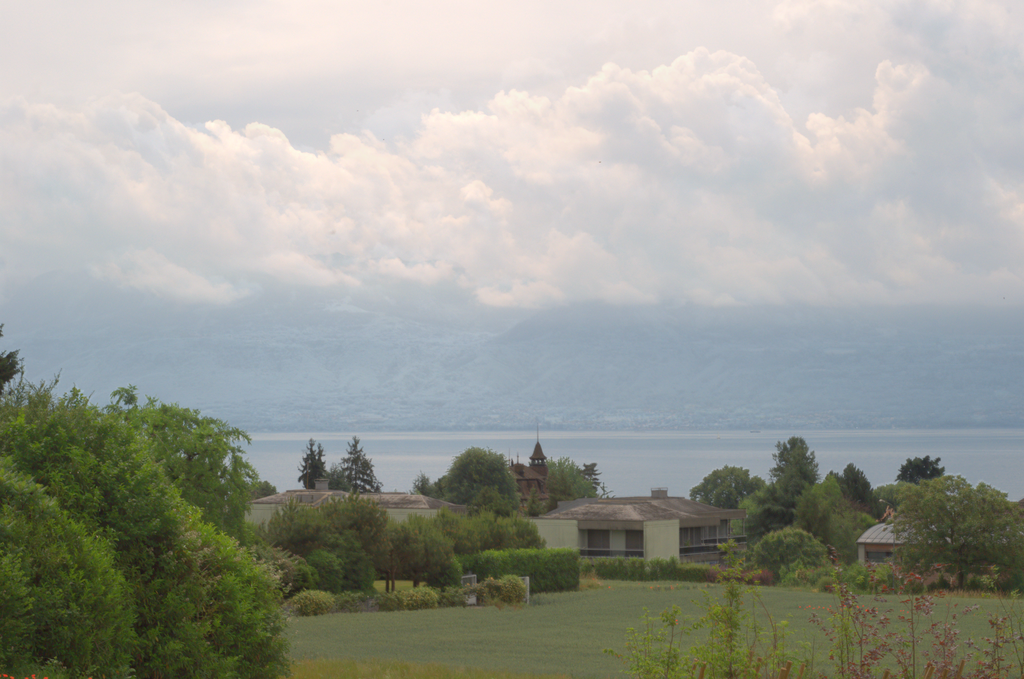}
	\includegraphics[width=50px]{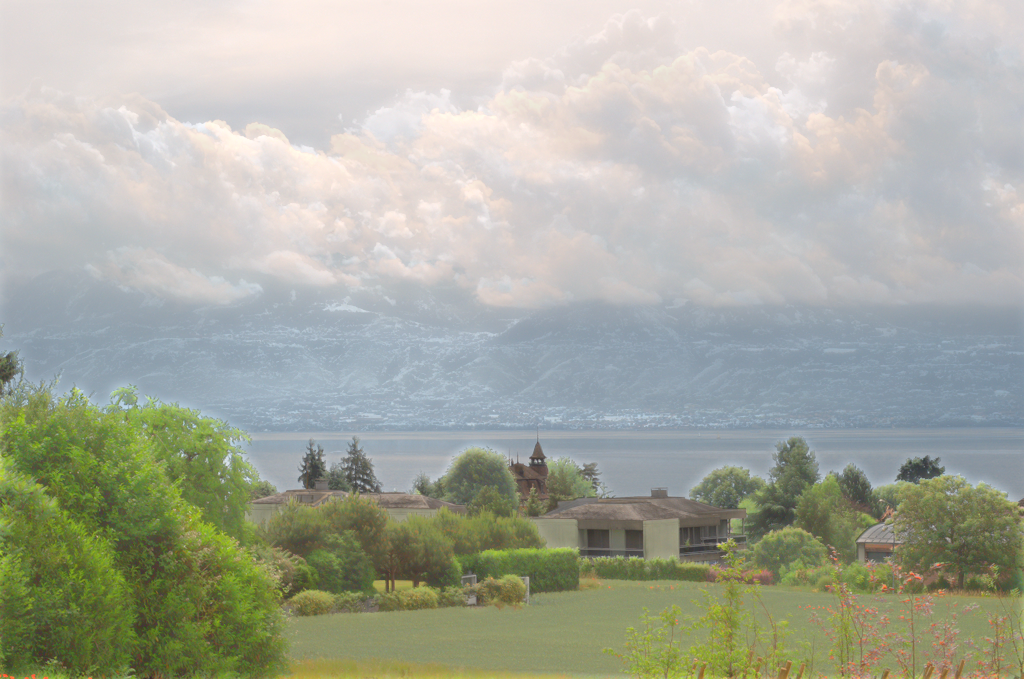}\\ [0.1cm]
	\includegraphics[width=50px]{43_fusion.png}
	\includegraphics[width=50px]{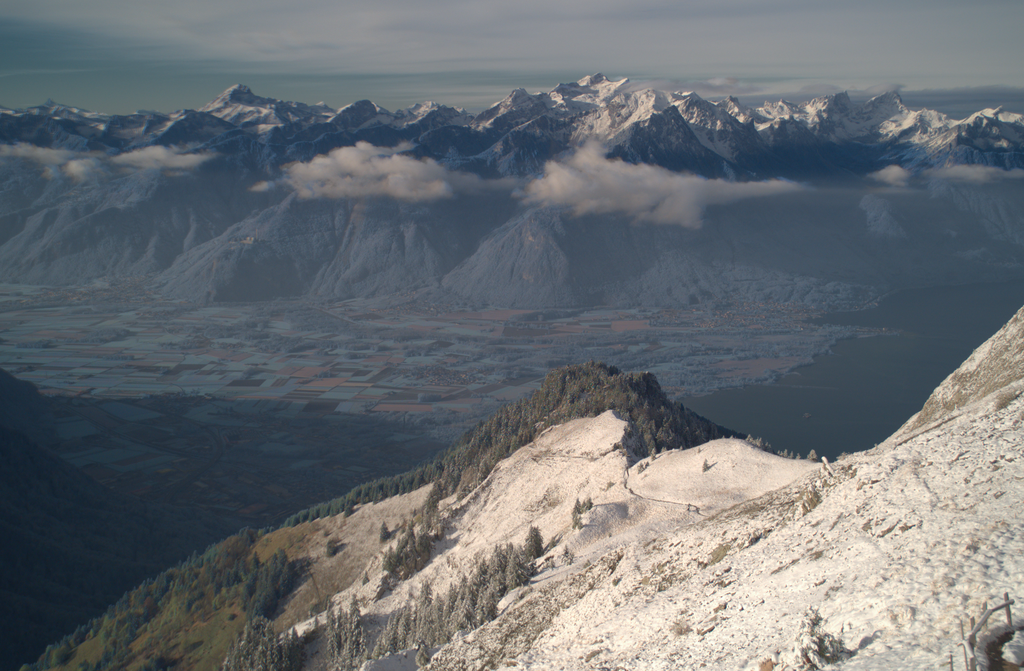}
	\includegraphics[width=50px]{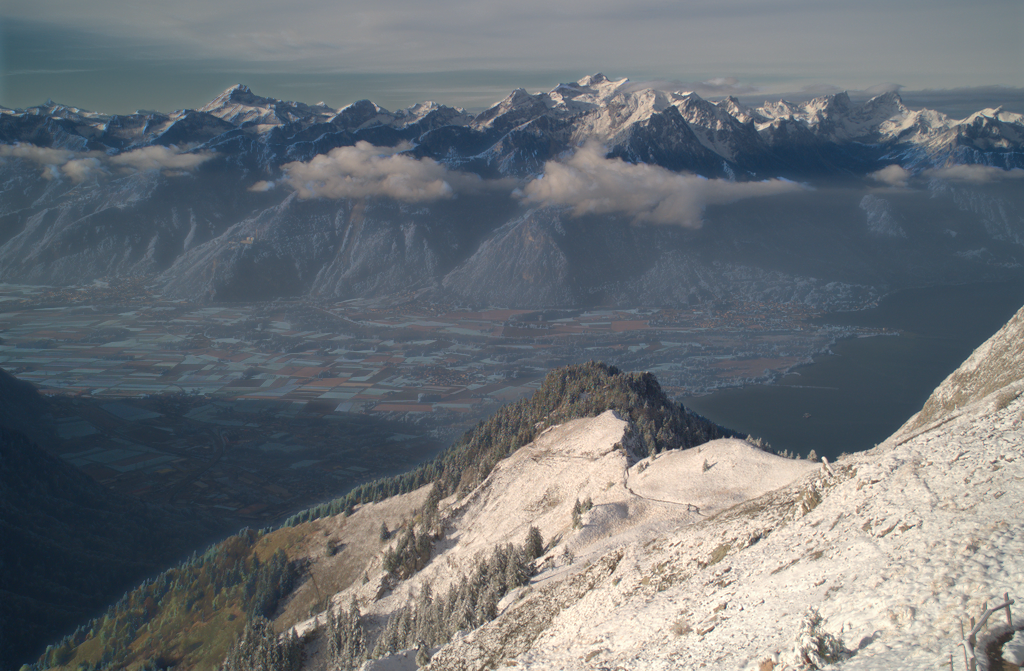}\\ [0.1cm]
	\includegraphics[width=50px]{51_fusion.png}
	\includegraphics[width=50px]{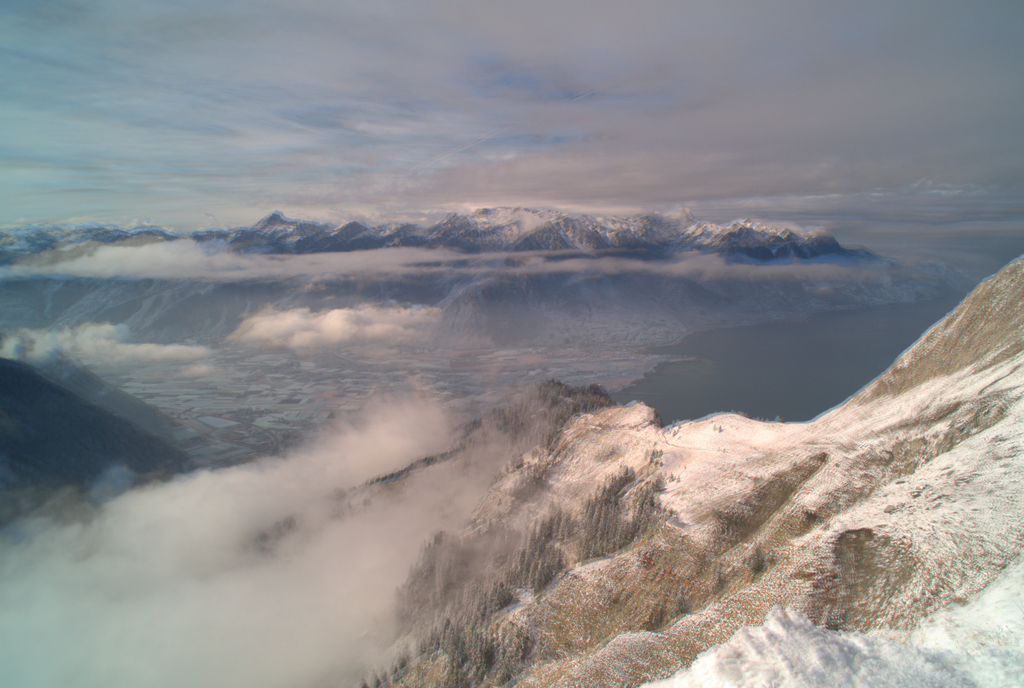}
	\includegraphics[width=50px]{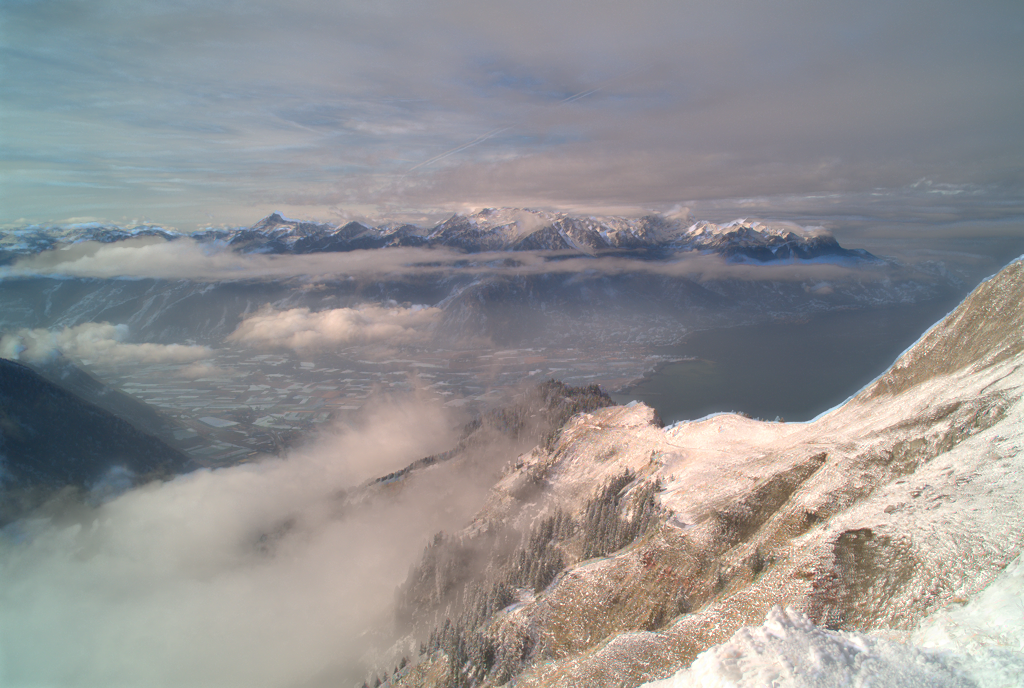}\\ [0.1cm]
	\includegraphics[width=50px]{55_fusion.png}
	\includegraphics[width=50px]{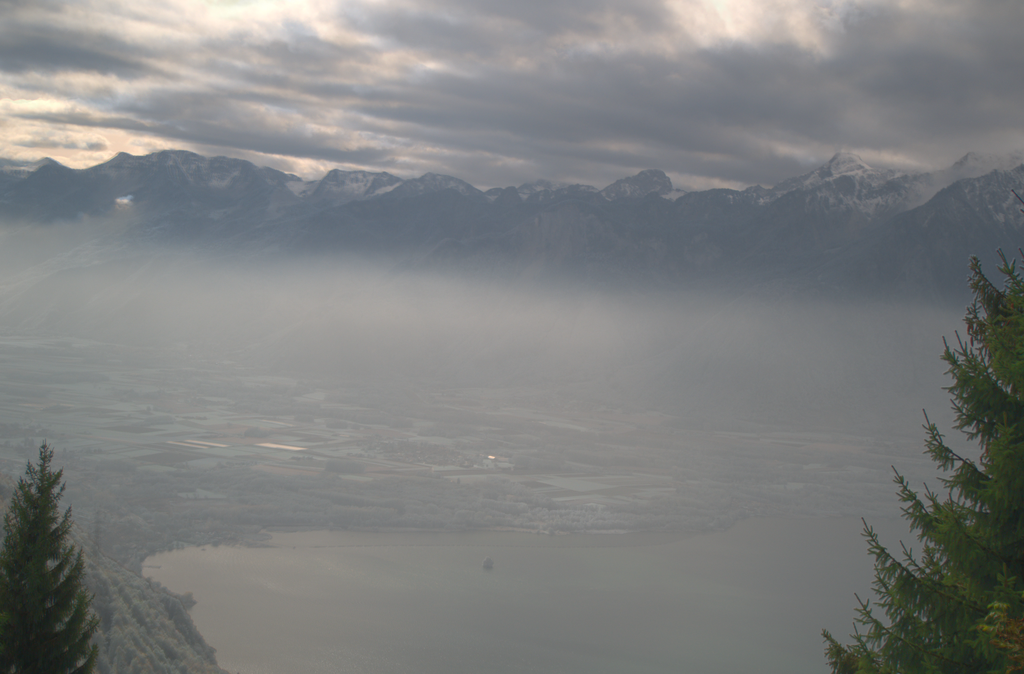}
	\includegraphics[width=50px]{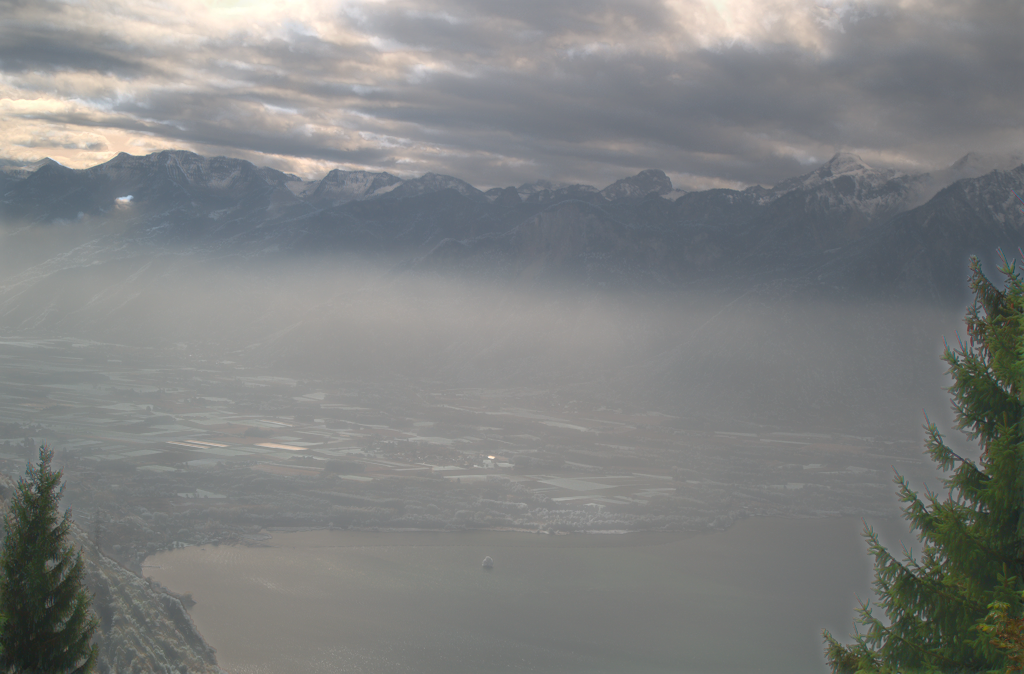}\\ [0.1cm]
	\includegraphics[width=50px]{35_fusion.png}
	\includegraphics[width=50px]{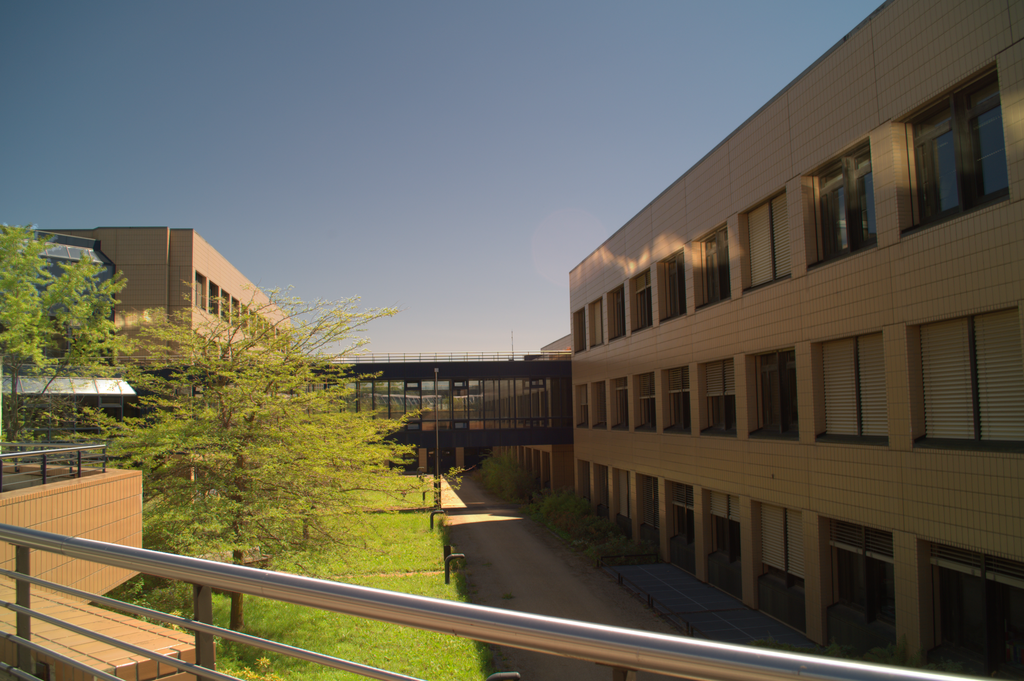}
	\includegraphics[width=50px]{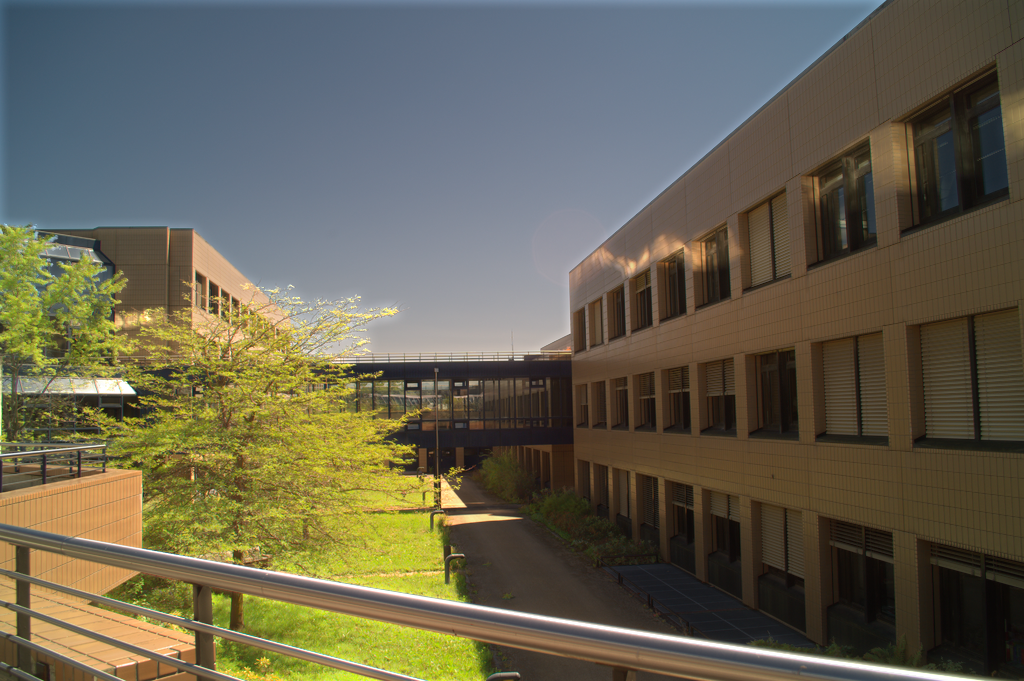}
	\caption{Comparison of fused images produced by the proposed approach (Left), global PCA fusion \cite{he2010multimodal} (Middle) and spectral fusion (Right).}
	\label{fig:fusion_pca}
\end{figure}

\begin{table}[htb]
	\centering
	\begin{tabular}{| l | c | c| c |}
        \hline
		Category & SuperPixel & PCA & Spectral \\
		\hline
		Country & \textbf{54.7} & 53.1 & 51.3 \\
		\hline
		Mountain & \textbf{58.4} & 56.6 & 54.4 \\
		\hline
        Urban & \textbf{76.4} & \textbf{76.4} & 74.5 \\
		\hline
        Street & 59.3 & \textbf{59.7} & 55.9 \\
		\hline
	\end{tabular}
	\caption{Percentage of initial Canny edges remaining in the fused images. The proposed super-pixel method better preserves edges than the other classic approaches for image fusion whatever the image category \cite{BS11}.}
	\label{table:metric}
\end{table}

\begin{table}[htb]
	\centering
	\begin{tabular}{| l | c | c| c |}
        \hline
		Category & SuperPixel & PCA & Spectral \\
		\hline
		Country & \textbf{81.5} & 74.1 & 78.9 \\
		\hline
		Mountain & \textbf{89.9} & 89.6 & 88.1 \\
		\hline
        Urban & 93.8 & \textbf{93.9} & 92.7 \\
		\hline
        Street & \textbf{87.5} & 87.3 & 85.5 \\
		\hline
	\end{tabular}
	\caption{Structure of similarity scores of the fused images.}
	\label{table:metricSSIM}
\end{table}

Figure \ref{fig:fusion_results} shows examples of multispectral images processed by the proposed fusion approach from the country, mountain, and urban image categories.
It reveals that the fused images contain details invisible in the RGB channels due to obstacles formed by haze, fog, or clouds. Moreover, they do not show high-frequency artifacts. 
It is important to note that although fusion masks are based on local patches, they cannot be detected on the fused images. Figure \ref{fig:fusion_pca} compares the local approach to global PCA fusion and spectral fusion. The proposed fusion is more informative and combines information from both spectral channels. The far mountains can be seen only by this method, and not by the PCA approach nor the spectral fusion. Although the spectral approach also emphasized details, it is performed globally and not locally like the proposed method.
This qualitative assessment suggests that local weighting of pixel fusion enhances fusion quality. 

Table \ref{table:metric} shows that the proposed method achieves the highest percentage of Canny \cite{canny1986computational} edge preservation in the fused images whatever their category. This provides evidence that the super-pixel approach better maintains details from the input multispectral images.
Finally, Table \ref{table:metricSSIM} reports the corresponding SSIM \cite{ssim, sara2019image} scores that are known to be correlated with the human visual system. It shows that the superpixel segmentation approach achieves the top SSIM. 


Both qualitative and quantitative results support the claim that the proposed method produces multispectral fused images that enhance visual information.

\section{Conclusions} \label{sec:conclusions}
We introduced a new method for multispectral fusion which applies a spatial soft map based on input image superpixel segmentation. This method shows advantages over existing approaches such that the details in the input images are preserved better in the fusion result, and still, the information of the color remains valid. In addition, we explained the theory behind the PCA and spectral fusion techniques and compared their principles to the proposed approach. As a whole, this paper produces an informative research work on the interesting problem of multispectral image fusion, in the color RGB to NIR domain.

	{\small
		\bibliographystyle{ieee}
		\bibliography{egbib}

\begin{thebibliography}{10}\itemsep=-1pt

\bibitem{achanta2012slic}
R.~Achanta, A.~Shaji, K.~Smith, A.~Lucchi, P.~Fua, and S.~S{\"u}sstrunk.
\newblock Slic superpixels compared to state-of-the-art superpixel methods.
\newblock {\em IEEE transactions on pattern analysis and machine intelligence},
  34(11):2274--2282, 2012.

\bibitem{agrawal2002formulation}
O.~P. Agrawal.
\newblock Formulation of euler--lagrange equations for fractional variational
  problems.
\newblock {\em Journal of Mathematical Analysis and Applications},
  272(1):368--379, 2002.

\bibitem{BS11}
M.~Brown and S.~S\"usstrunk.
\newblock Multispectral {SIFT} for scene category recognition.
\newblock In {\em Computer Vision and Pattern Recognition (CVPR11)}, pages
  177--184, Colorado Springs, June 2011.

\bibitem{burt1987laplacian}
P.~J. Burt and E.~H. Adelson.
\newblock The laplacian pyramid as a compact image code.
\newblock In {\em Readings in computer vision}, pages 671--679. Elsevier, 1987.

\bibitem{canny1986computational}
J.~Canny.
\newblock A computational approach to edge detection.
\newblock {\em IEEE Transactions on pattern analysis and machine intelligence},
  (6):679--698, 1986.

\bibitem{chen2017deep}
Y.~Chen, C.~Li, P.~Ghamisi, X.~Jia, and Y.~Gu.
\newblock Deep fusion of remote sensing data for accurate classification.
\newblock {\em IEEE Geoscience and Remote Sensing Letters}, 14(8):1253--1257,
  2017.

\bibitem{chipman1995wavelets}
L.~J. Chipman, T.~M. Orr, and L.~N. Graham.
\newblock Wavelets and image fusion.
\newblock In {\em Proceedings., International Conference on Image Processing},
  volume~3, pages 248--251. IEEE, 1995.

\bibitem{guo2020rsdehazenet}
J.~Guo, J.~Yang, H.~Yue, H.~Tan, C.~Hou, and K.~Li.
\newblock Rsdehazenet: Dehazing network with channel refinement for
  multispectral remote sensing images.
\newblock {\em IEEE Transactions on geoscience and remote sensing},
  59(3):2535--2549, 2020.

\bibitem{he2010multimodal}
C.~He, Q.~Liu, H.~Li, and H.~Wang.
\newblock Multimodal medical image fusion based on ihs and pca.
\newblock {\em Procedia Engineering}, 7:280--285, 2010.

\bibitem{horn1981determining}
B.~K. Horn and B.~G. Schunck.
\newblock Determining optical flow.
\newblock {\em Artificial intelligence}, 17(1-3):185--203, 1981.

\bibitem{james2014medical}
A.~P. James and B.~V. Dasarathy.
\newblock Medical image fusion: A survey of the state of the art.
\newblock {\em Information fusion}, 19:4--19, 2014.

\bibitem{jia2019collaborative}
S.~Jia, X.~Deng, J.~Zhu, M.~Xu, J.~Zhou, and X.~Jia.
\newblock Collaborative representation-based multiscale superpixel fusion for
  hyperspectral image classification.
\newblock {\em IEEE Transactions on Geoscience and Remote Sensing},
  57(10):7770--7784, 2019.

\bibitem{kumar2006pca}
S.~S. Kumar and S.~Muttan.
\newblock Pca-based image fusion.
\newblock In {\em Algorithms and Technologies for Multispectral, Hyperspectral,
  and Ultraspectral Imagery XII}, volume 6233, page 62331T. International
  Society for Optics and Photonics, 2006.

\bibitem{li2013image}
S.~Li, X.~Kang, and J.~Hu.
\newblock Image fusion with guided filtering.
\newblock {\em IEEE Transactions on Image processing}, 22(7):2864--2875, 2013.

\bibitem{lowe2004distinctive}
D.~G. Lowe.
\newblock Distinctive image features from scale-invariant keypoints.
\newblock {\em International journal of computer vision}, 60(2):91--110, 2004.

\bibitem{ofir2021classic}
N.~Ofir and J.-C. Nebel.
\newblock Classic versus deep approaches to address computer vision challenges.
\newblock {\em arXiv preprint arXiv:2101.09744}, 2021.

\bibitem{ofir2018deep}
N.~Ofir, S.~Silberstein, H.~Levi, D.~Rozenbaum, Y.~Keller, and S.~D. Bar.
\newblock Deep multi-spectral registration using invariant descriptor learning.
\newblock In {\em 2018 25th IEEE International Conference on Image Processing
  (ICIP)}, pages 1238--1242. IEEE, 2018.

\bibitem{ofir2018registration}
N.~Ofir, S.~Silberstein, D.~Rozenbaum, Y.~Keller, and S.~D. Bar.
\newblock Registration and fusion of multi-spectral images using a novel edge
  descriptor.
\newblock In {\em 2018 25th IEEE International Conference on Image Processing
  (ICIP)}, pages 1857--1861. IEEE, 2018.

\bibitem{ofir2021classicThesis}
Y.~N. Ofir.
\newblock {\em Classic versus deep learning approaches to address computer
  vision challenges: a study of faint edge detection and multispectral image
  registration}.
\newblock PhD thesis, Kingston University, 2021.

\bibitem{pajares2004wavelet}
G.~Pajares and J.~M. De~La~Cruz.
\newblock A wavelet-based image fusion tutorial.
\newblock {\em Pattern recognition}, 37(9):1855--1872, 2004.

\bibitem{sara2019image}
U.~Sara, M.~Akter, and M.~S. Uddin.
\newblock Image quality assessment through fsim, ssim, mse and psnr—a
  comparative study.
\newblock {\em Journal of Computer and Communications}, 7(3):8--18, 2019.

\bibitem{srivastava2017combination}
A.~Srivastava, V.~Bhateja, and A.~Moin.
\newblock Combination of pca and contourlets for multispectral image fusion.
\newblock In {\em Proceedings of the international conference on data
  engineering and communication technology}, pages 577--585. Springer, 2017.

\bibitem{takumi2017multispectral}
K.~Takumi, K.~Watanabe, Q.~Ha, A.~Tejero-De-Pablos, Y.~Ushiku, and T.~Harada.
\newblock Multispectral object detection for autonomous vehicles.
\newblock In {\em Proceedings of the on Thematic Workshops of ACM Multimedia
  2017}, pages 35--43, 2017.

\bibitem{toet1990hierarchical}
A.~Toet.
\newblock Hierarchical image fusion.
\newblock {\em Machine Vision and Applications}, 3(1):1--11, 1990.

\bibitem{wagner2016multispectral}
J.~Wagner, V.~Fischer, M.~Herman, S.~Behnke, et~al.
\newblock Multispectral pedestrian detection using deep fusion convolutional
  neural networks.
\newblock In {\em ESANN}, volume 587, pages 509--514, 2016.

\bibitem{wang2009mean}
Z.~Wang and A.~C. Bovik.
\newblock Mean squared error: Love it or leave it? a new look at signal
  fidelity measures.
\newblock {\em IEEE signal processing magazine}, 26(1):98--117, 2009.

\bibitem{ssim}
Z.~Wang, A.~C. Bovik, H.~R. Sheikh, and E.~P. Simoncelli.
\newblock Image quality assessment: from error visibility to structural
  similarity.
\newblock {\em IEEE transactions on image processing}, 13(4):600--612, 2004.

\bibitem{wei2015hyperspectral}
Q.~Wei, J.~Bioucas-Dias, N.~Dobigeon, and J.-Y. Tourneret.
\newblock Hyperspectral and multispectral image fusion based on a sparse
  representation.
\newblock {\em IEEE Transactions on Geoscience and Remote Sensing},
  53(7):3658--3668, 2015.

\bibitem{zeng2010fusion}
Y.~Zeng, W.~Huang, M.~Liu, H.~Zhang, and B.~Zou.
\newblock Fusion of satellite images in urban area: Assessing the quality of
  resulting images.
\newblock In {\em 2010 18th international conference on geoinformatics}, pages
  1--4. IEEE, 2010.

\bibitem{zhao2018multi}
W.~Zhao, D.~Wang, and H.~Lu.
\newblock Multi-focus image fusion with a natural enhancement via a joint
  multi-level deeply supervised convolutional neural network.
\newblock {\em IEEE Transactions on Circuits and Systems for Video Technology},
  29(4):1102--1115, 2018.

\end{thebibliography}
	}
	
\end{document}